\pdfoutput=1
\documentclass{article}

\usepackage{times}
\usepackage{graphicx} % more modern
\usepackage{subfigure} 

\usepackage{natbib}

\usepackage{algorithm}
\usepackage{algorithmic}

\usepackage{hyperref}

\usepackage[accepted]{icml2016}

\icmltitlerunning{Exploring the Limits of Language Modeling}

\begin{document} 

\twocolumn[
\icmltitle{Exploring the Limits of Language Modeling}

% It is OKAY to include author information, even for blind
% submissions: the style file will automatically remove it for you
% unless you've provided the [accepted] option to the icml2016
% package.
\icmlauthor{Rafal Jozefowicz}{rafalj@google.com}
\icmlauthor{Oriol Vinyals}{vinyals@google.com}
\icmlauthor{Mike Schuster}{schuster@google.com}
\icmlauthor{Noam Shazeer}{noam@google.com}
\icmlauthor{Yonghui Wu}{yonghui@google.com}
\icmladdress{\vspace{0.2cm} Google Brain}

% You may provide any keywords that you 
% find helpful for describing your paper; these are used to populate 
% the "keywords" metadata in the PDF but will not be shown in the document
% \icmlkeywords{boring formatting information, machine learning, ICML}

\vskip 0.3in
]

\begin{abstract} 
In this work we explore recent advances in Recurrent Neural Networks for large scale Language Modeling, a task central to language understanding. We extend current models to deal with two key challenges present in this task: corpora and vocabulary sizes, and complex, long term structure of language. We perform an exhaustive study on techniques such as character Convolutional Neural Networks or Long-Short Term Memory, on the One Billion Word Benchmark. Our best single model significantly improves state-of-the-art perplexity from 51.3 down to 30.0 (whilst reducing the number of parameters by a factor of 20), while an ensemble of models sets a new record by improving perplexity from 41.0 down to 23.7. We also release these models for the NLP and ML community to study and improve upon.
\end{abstract} 

\section{Introduction}
\label{intro}

Language Modeling (LM) is a task central to Natural Language Processing (NLP) and Language Understanding. Models which can accurately place distributions over sentences not only encode complexities of language such as grammatical structure, but also distill a fair amount of information about the knowledge that a corpora may contain. Indeed, models that are able to assign a low probability to sentences that are grammatically correct but unlikely may help other tasks in fundamental language understanding like question answering, machine translation, or text summarization.

LMs have played a key role in traditional NLP tasks such as speech recognition \cite{mikolov2010recurrent, arisoy2012deep}, machine translation \cite{schwenk2012large, vaswani2013decoding}, or text summarization \cite{rush2015neural,filippova2015sentence}. Often (although not always), training better language models improves the underlying metrics of the downstream task (such as word error rate for speech recognition, or BLEU score for translation), which makes the task of training better LMs valuable by itself.

Further, when trained on vast amounts of data, language models compactly extract knowledge encoded in the training data. For example, when trained on movie subtitles \cite{serban2015,vinyals2015neural}, these language models are able to generate basic answers to questions about object colors, facts about people, etc. Lastly, recently proposed sequence-to-sequence models employ conditional language models \cite{mikolov2012context} as their key component to solve diverse tasks like machine translation \cite{sutskever2014sequence, cho2014learning, kalchbrenner2014convolutional} or video generation \cite{srivastava2015unsupervised}.

\begin{figure}[t]
\vskip 0.2in
\begin{center}
\centerline{\includegraphics[width=\columnwidth]{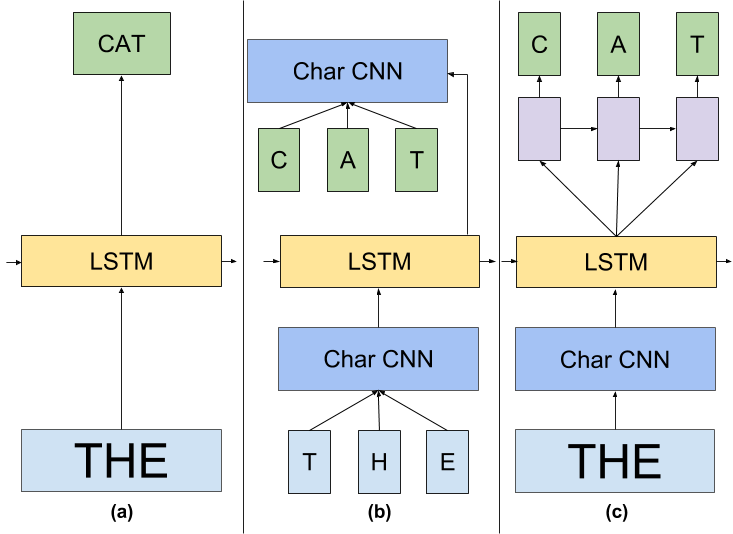}}
\caption{A high-level diagram of the models presented in this paper. (a) is a standard LSTM LM. (b) represents an LM where both input and Softmax embeddings have been replaced by a character CNN. In (c) we replace the Softmax by a next character prediction LSTM network.}
\label{main_figure}
\end{center}
\vspace{-10pt}
\end{figure} 

Deep Learning and Recurrent Neural Networks (RNNs) have fueled language modeling research in the past years as it allowed researchers to explore many tasks for which the strong conditional independence assumptions are unrealistic. Despite the fact that simpler models, such as N-grams, only use a short history of previous words to predict the next word, they are still a key component to high quality, low perplexity LMs. Indeed, most recent work on large scale LM has shown that RNNs are great in combination with N-grams, as they may have different strengths that complement N-gram models, but worse when considered in isolation \cite{mikolov2011empirical, mikolov2012statistical, chelba2013one,williams2015scaling,Blackout,shazeer2015sparse}.

We believe that, despite much work being devoted to small data sets like the Penn Tree Bank (PTB) \cite{marcus1993building}, research on larger tasks is very relevant as overfitting is not the main limitation in current language modeling, but is the main characteristic of the PTB task. Results on larger corpora usually show better what matters as many ideas work well on small data sets but fail to improve on larger data sets. Further, given current hardware trends and vast amounts of text available on the Web, it is much more straightforward to tackle large scale modeling than it used to be. Thus, we hope that our work will help and motivate researchers to work on traditional LM beyond PTB -- for this purpose, we will open-source our models and training recipes. 

We focused on a well known, large scale LM benchmark: the One Billion Word Benchmark data set \cite{chelba2013one}. This data set is much larger than PTB (one thousand fold, ~800k word vocabulary and ~1B words training data) and far more challenging. Similar to Imagenet \cite{deng2009imagenet}, which helped advance computer vision, we believe that releasing and working on large data sets and models with clear benchmarks will help advance Language Modeling.

The contributions of our work are as follows:

\begin{itemize}
\item We explored, extended and tried to unify some of the current research on large scale LM.
\item Specifically, we designed a Softmax loss which is based on character level CNNs, is efficient to train, and is as precise as a full Softmax which has orders of magnitude more parameters.
\item Our study yielded significant improvements to the state-of-the-art on a well known, large scale LM task: from 51.3 down to 30.0 perplexity for single models whilst reducing the number of parameters by a factor of 20.
\item We show that an ensemble of a number of different models can bring down perplexity on this task to 23.7, a large improvement compared to current state-of-art.
\item We share the model and recipes in order to help and motivate further research in this area.
\end{itemize}

In Section~\ref{relwork} we review important concepts and previous work on language modeling. Section~\ref{model} presents our contributions to the field of neural language modeling, emphasizing large scale recurrent neural network training. Sections \ref{exps} and \ref{results} aim at exhaustively describing our experience and understanding throughout the project, as well as emplacing our work relative to other known approaches.

\section{Related Work}
\label{relwork}

In this section we describe previous work relevant to the approaches discussed in this paper. A more detailed discussion on language modeling research is provided in \cite{mikolov2012statistical}.

\subsection{Language Models}
\label{lms}

Language Modeling (LM) has been a central task in NLP. The goal of LM is to learn a probability distribution over sequences of symbols pertaining to a language. Much work has been done on both parametric (e.g., log-linear models) and non-parametric approaches (e.g., count-based LMs). Count-based approaches (based on statistics of N-grams) typically add smoothing which account for unseen (yet possible) sequences, and have been quite successful. To this extent, Kneser-Ney smoothed 5-gram models \cite{kneser1995improved} are a fairly strong baseline which, for large amounts of training data, have challenged other parametric approaches based on Neural Networks \cite{bengio2006neural}.

Most of our work is based on Recurrent Neural Networks (RNN) models which retain long term dependencies. To this extent, we used the Long-Short Term Memory model \cite{hochreiter1997long} which uses a gating mechanism \cite{gers2000learning} to ensure proper propagation of information through many time steps. Much work has been done on small and large scale RNN-based LMs \cite{mikolov2010recurrent,  mikolov2012statistical, chelba2013one,zaremba2014recurrent, williams2015scaling,Blackout,wang2015larger,ji2015document}. The architectures that we considered in this paper are represented in Figure~\ref{main_figure}.

In our work, we train models on the popular One Billion Word Benchmark, which can be considered to be a medium-sized data set for count-based LMs but a very large data set for NN-based LMs. This regime is most interesting to us as we believe learning a very good model of human language is a complex task which will require large models, and thus large amounts of data. Further advances in data availability and computational resources helped our study. We argue this leap in scale enabled tremendous advances in deep learning. A clear example found in computer vision is Imagenet \cite{deng2009imagenet}, which enabled learning complex vision models from large amounts of data \cite{krizhevsky2012imagenet}.

A crucial aspect which we discuss in detail in later sections is the size of our models. Despite the large number of parameters, we try to minimize computation as much as possible by adopting a strategy proposed in \cite{sak2014long} of projecting a relatively big recurrent state space down so that the matrices involved remain relatively small, yet the model has large memory capacity.

\subsection{Convolutional Embedding Models}
\label{input_cnn}

There is an increased interest in incorporating character-level inputs to build word embeddings for various NLP problems, including part-of-speech tagging, parsing and language modeling \cite{ling2015finding, kim2015character, ballesteros2015improved}. The additional character information has been shown useful on relatively small benchmark data sets.

The approach proposed in \cite{ling2015finding} builds word embeddings using bidirectional LSTMs \cite{schuster1997brnn,graves2005framewise} over the characters. The recurrent networks process sequences of characters from both sides and their final state vectors are concatenated. The resulting representation is then fed to a Neural Network. This model achieved very good results on a part-of-speech tagging task.

In \cite{kim2015character}, the words characters are processed by a 1-d CNN \cite{le1990handwritten} with max-pooling across the sequence for each convolutional feature. The resulting features are fed to a 2-layer highway network \cite{srivastava2015training}, which allows the embedding to learn semantic representations. The model was evaluated on small-scale language modeling experiments for various languages and matched the best results on the PTB data set despite having 60\% fewer parameters.

\subsection{Softmax Over Large Vocabularies}
\label{large_soft}

Assigning probability distributions over large vocabularies is computationally challenging. For modeling language, maximizing log-likelihood of a given word sequence leads to optimizing cross-entropy between the target probability distribution (e.g., the target word we should be predicting), and our model predictions $p$. Generally, predictions come from a linear layer followed by a Softmax non-linearity: $p(w) = \frac{\exp(z_w)}{\sum_{w' \in V} \exp(z_{w'})}$ where $z_w$ is the logit corresponding to a word $w$. The logit is generally computed as an inner product $z_w = h^Te_w$ where $h$ is a context vector and $e_w$ is a ``word embedding'' for $w$.

The main challenge when $|V|$ is very large (in the order of one million in this paper) is the fact that computing all inner products between $h$ and all embeddings becomes prohibitively slow during training (even when exploiting matrix-matrix multiplications and modern GPUs). Several approaches have been proposed to cope with the scaling issue: importance sampling \cite{bengio2003quick, bengio2008adaptive}, Noise Contrastive Estimation (NCE) \cite{gutmann2010noise, mnih2013learning}, self normalizing partition functions \cite{vincent2015efficient} or Hierarchical Softmax \cite{morin2005hierarchical, mnih2009scalable} -- they all offer good solutions to this problem. We found importance sampling to be quite effective on this task, and explain the connection between it and NCE in the following section, as they are closely related.

\section{Language Modeling Improvements}
\label{model}

Recurrent Neural Networks based LMs employ the chain rule to model joint probabilities over word sequences:

$$p(w_1,\ldots,w_N) = \prod_{i=1}^N p(w_i | w_1, \ldots, w_{i-1})$$
where the context of all previous words is encoded with an LSTM, and the probability over words uses a Softmax (see Figure~\ref{main_figure}(a)).

\subsection{Relationship between Noise Contrastive Estimation and Importance Sampling}
\label{is_soft}

As discussed in Section~\ref{large_soft}, a large scale Softmax is necessary for training good LMs because of the vocabulary size. A Hierarchical Softmax \cite{mnih2009scalable} employs a tree in which the probability distribution over words is decomposed into a product of two probabilities for each word, greatly reducing training and inference time as only the path specified by the hierarchy needs to be computed and updated. Choosing a good hierarchy is important for obtaining good results and we did not explore this approach further for this paper as sampling methods worked well for our setup.

Sampling approaches are only useful during training, as they propose an approximation to the loss which is cheap to compute (also in a distributed setting) -- however, at inference time one still has to compute the normalization term over all words. Noise Contrastive Estimation (NCE) proposes to consider a surrogate binary classification task in which a classifier is trained to discriminate between true data, or samples coming from some arbitrary distribution. If both the noise and data distributions were known, the optimal classifier would be:

$$p(Y=true | w) = \frac{p_d(w)}{p_d(w) + k p_n(w)}$$
where $Y$ is the binary random variable indicating whether $w$ comes from the true data distribution, $k$ is the number of negative samples per positive word, and $p_d$ and $p_n$ are the data and noise distribution respectively (we dropped any dependency on previous words for notational simplicity).

It is easy to show that if we train a logistic classifier $p_\theta(Y=true|w) = \sigma(s_\theta(w,h)-\log k p_n(w))$ where $\sigma$ is the logistic function, then, $p'(w)=softmax(s_\theta(w,h))$ is a good approximation of $p_d(w)$ ($s_\theta$ is a logit which e.g. an LSTM LM computes).

The other technique, which is based on importance sampling (IS), proposes to directly approximate the partition function (which comprises a sum over all words) with an estimate of it through importance sampling. Though the methods look superficially similar, we will derive a similar surrogate classification task akin to NCE which arrives at IS, showing a strong connection between the two.

Suppose that, instead of having a binary task to decide if a word comes from the data or from the noise distribution, we want to identify the words coming from the true data distribution in a set $W=\{w_1,\ldots,w_{k+1}\}$, comprised of $k$ noise samples and one data distribution sample. Thus, we can train a multiclass loss over a multinomial random variable $Y$ which maximizes $\log p(Y = 1 | W)$, assuming w.l.o.g. that $w_1 \in W$ is always the word coming from true data. By Bayes rule, and ignoring terms that are constant with respect to $Y$, we can write:

$$p(Y=k | W) \propto_Y \frac{p_d(w_k)}{p_n(w_k)} $$
and, following a similar argument than for NCE, if we define $p(Y=k | W)=softmax(s_\theta(w_k)-\log  p_n(w_k))$ then $p'(w)=softmax(s_\theta(w,h))$ is a good approximation of $p_d(word)$. Note that the only difference between NCE and IS is that, in NCE, we define a binary classification task between true or noise words with a logistic loss, whereas in IS we define a multiclass classification problem with a Softmax and cross entropy loss. We hope that our derivation helps clarify the similarities and differences between the two. In particular, we observe that IS, as it optimizes a multiclass classification task (in contrast to solving a binary task), may be a better choice. Indeed, the updates to the logits with IS are tied whereas in NCE they are independent.

\subsection{CNN Softmax}
\label{cnn_softmax}

The character-level features allow for a smoother and compact parametrization of the word embeddings. Recent efforts on small scale language modeling have used CNN character embeddings for the input embeddings \cite{kim2015character}. Although not as straightforward, we propose an extension to this idea to also reduce the number of parameters of the Softmax layer. Recall from Section~\ref{large_soft} that the Softmax computes a logit as $z_w=h^Te_w$ where $h$ is a context vector and $e_w$ the word embedding. Instead of building a matrix of $|V|\times |h|$ (whose rows correspond to $e_w$), we produce $e_w$ with a CNN over the characters of $w$ as $e_w = CNN(chars_w)$ -- we call this a CNN Softmax. We used the same network architecture to dynamically generate the Softmax word embeddings without sharing the parameters with the input word-embedding sub-network. For inference, the vectors $e_w$ can be precomputed, so there is no computational complexity increase w.r.t. the regular Softmax.

We note that, when using an importance sampling loss such as the one described in Section~\ref{is_soft}, only a few logits have non-zero gradient (those corresponding to the true and sampled words). With a Softmax where $e_w$ are independently learned word embeddings, this is not a problem. But we observed that, when using a CNN, all the logits become tied as the function mapping from $w$ to $e_w$ is quite smooth. As a result, a much smaller learning rate had to be used. Even with this, the model lacks capacity to differentiate between words that have very different meanings but that are spelled similarly. Thus, a reasonable compromise was to add a small correction factor which is learned per word, such that:

$$z_w=h^TCNN(chars_w) + h^T M corr_w$$
where $M$ is a matrix projecting a low-dimensional embedding vector $corr_w$ back up to the dimensionality of the projected LSTM hidden state of $h$. This amounts to adding a bottleneck linear layer, and brings the CNN Softmax much closer to our best result, as can be seen in Table~\ref{main-results-table}, where adding a 128-dim correction halves the gap between regular and the CNN Softmax.

Aside from a big reduction in the number of parameters and incorporating morphological knowledge from words, the other benefit of this approach is that out-of-vocabulary (OOV) words can easily be scored. This may be useful for other problems such as Machine Translation where handling out-of-vocabulary words is very important \cite{luong2014addressing}. This approach also allows parallel training over various data sets since the model is no longer explicitly parametrized by the vocabulary size -- or the language. This has shown to help when using byte-level input embeddings for named entity recognition \cite{gillick2015multilingual}, and we hope it will enable similar gains when used to map onto words.

\begin{table*}[t!]
\caption{Best results of single models on the 1B word benchmark. Our results are shown below previous work.}
\label{main-results-table}
\vskip 0.15in
\begin{center}
\begin{small}
\begin{sc}
\begin{tabular}{lcr}
\hline
\abovespace\belowspace
Model & Test Perplexity & Number of Params [billions] \\
\hline
\abovespace
Sigmoid-RNN-2048 \cite{Blackout} & 68.3 & 4.1 \\
Interpolated KN 5-gram, 1.1B n-grams \cite{chelba2013one} & 67.6 & 1.76 \\
Sparse Non-Negative Matrix LM \cite{shazeer2015sparse} & 52.9 & 33 \\
RNN-1024 + MaxEnt 9-gram features \cite{chelba2013one} & 51.3 & 20 \\
\hline
\abovespace
LSTM-512-512 & 54.1 & 0.82 \\
LSTM-1024-512 & 48.2 & 0.82 \\
LSTM-2048-512 & 43.7 & 0.83 \\
LSTM-8192-2048 (No Dropout) & 37.9 & 3.3 \\
LSTM-8192-2048 (50\% Dropout) & 32.2 & 3.3 \\
2-Layer LSTM-8192-1024 (BIG LSTM) & 30.6 & 1.8 \\
BIG LSTM+CNN Inputs & \textbf{30.0} & \textbf{1.04} \\
\abovespace
BIG LSTM+CNN Inputs + CNN Softmax & 39.8 & \textbf{0.29} \\
BIG LSTM+CNN Inputs + CNN Softmax + 128-dim correction & 35.8 & \textbf{0.39} \\
BIG LSTM+CNN Inputs + Char LSTM predictions & 47.9 & \textbf{0.23} \\
\hline
\end{tabular}
\end{sc}
\end{small}
\end{center}
\vskip -0.1in
\end{table*}

\subsection{Char LSTM Predictions}
\label{charlstm}

The CNN Softmax layer can handle arbitrary words and is much more efficient in terms of number of parameters than the full Softmax matrix. It is, though, still considerably slow, as to evaluate perplexities we need to compute the partition function. A class of models that solve this problem more efficiently are character-level LSTMs \cite{sutskever2011generating, graves2013generating}. They make predictions one character at a time, thus allowing to compute probabilities over a much smaller vocabulary. On the other hand, these models are more difficult to train and seem to perform worse even in small tasks like PTB \cite{graves2013generating}. Most likely this is due to the sequences becoming much longer on average as the LSTM reads the input character by character instead of word by word.

Thus, we combine the word and character-level models by feeding a word-level LSTM hidden state $h$ into a small LSTM that predicts the target word one character at a time (see Figure~\ref{main_figure}(c)). In order to make the whole process reasonably efficient, we train the standard LSTM model until convergence, freeze its weights, and replace the standard word-level Softmax layer with the aforementioned character-level LSTM.

The resulting model scales independently of vocabulary size -- both for training and inference. However, it does seem to be worse than regular and CNN Softmax -- we are hopeful that further research will enable these models to replace fixed vocabulary models whilst being computationally attractive.

\section{Experiments}
\label{exps}

All experiments were run using the TensorFlow system \cite{tensorflow2015-whitepaper}, with the exception of some older models which were used in the ensemble.

\subsection{Data Set}
\label{data set}

The experiments are performed on the 1B Word Benchmark data set introduced by \cite{chelba2013one}, which is a publicly available benchmark for measuring progress of statistical language modeling. The data set contains about 0.8B words with a vocabulary of 793471 words, including sentence boundary markers. All the sentences are shuffled and the duplicates are removed. The words that are out of vocabulary (OOV) are marked with a special UNK token (there are approximately 0.3\% such words).

\subsection{Model Setup}
\label{experiments}

The typical measure used for reporting progress in language modeling is perplexity, which is the average per-word log-probability on the holdout data set: $e^{-\frac{1}{N}\sum_i \ln{p_{w_i}}}$. We follow the standard procedure and sum over all the words (including the end of sentence symbol).

We used the 1B Word Benchmark data set without any pre-processing. Given the shuffled sentences, they are input to the network as a batch of independent streams of words. Whenever a sentence ends, a new one starts without any padding (thus maximizing the occupancy per batch).

For the models that consume characters as inputs or as targets, each word is fed to the model as a sequence of character IDs of preespecified length (see Figure~\ref{main_figure}(b)). The words were processed to include special begin and end of word tokens and were padded to reach the expected length. I.e. if the maximum word length was 10, the word ``\texttt{cat}'' would be transformed to ``\texttt{\$cat\^{}               }'' due to the CNN model.

In our experiments we found that limiting the maximum word length in training to 50 was sufficient to reach very good results while 32 was clearly insufficient. We used 256 characters in our vocabulary and the non-ascii symbols were represented as a sequence of bytes.

\subsection{Model Architecture}

We evaluated many variations of RNN LM architectures. These include the dimensionalities of the embedding layers, the state, projection sizes, and number of LSTM layers to use. Exhaustively trying all combinations would be extremely time consuming for such a large data set, but our findings suggest that LSTMs with a projection layer (i.e., a bottleneck between hidden states as in \cite{sak2014long}) trained with truncated BPTT \cite{williams1990efficient} for 20 steps performed well. 

Following \cite{zaremba2014recurrent} we use dropout \cite{srivastava2013improving} before and after every LSTM layer. The biases of LSTM forget gate were initialized to 1.0 \cite{jozefowicz2015empirical}. The size of the models will be described in more detail in the following sections, and the choices of hyper-parameters will be released as open source upon publication.

For any model using character embedding CNNs, we closely follow the architecture from \cite{kim2015character}. The only important difference is that we use a larger number of convolutional features of 4096 to give enough capacity to the model. The resulting embedding is then linearly transformed to match the LSTM projection sizes. This allows it to match the performance of regular word embeddings but only uses a small fraction of parameters.

\begin{table*}[t]
\caption{Best results of ensembles on the 1B Word Benchmark.}
\label{ensemble-results-table}
\vskip 0.15in
\begin{center}
\begin{small}
\begin{sc}
\begin{tabular}{lr}
\hline
\abovespace\belowspace
Model & Test Perplexity \\
\hline
\abovespace
Large Ensemble \cite{chelba2013one} & 43.8 \\ 
RNN+KN-5 \cite{williams2015scaling} & 42.4 \\
RNN+KN-5 \cite{Blackout} & 42.0 \\
RNN+SNM10-skip \cite{shazeer2015sparse} & 41.3 \\
Large Ensemble \cite{shazeer2015sparse} & 41.0 \\
\hline
\abovespace
Our 10 best LSTM models (equal weights) & 26.3 \\
Our 10 best LSTM models (optimal weights) & 26.1 \\
10 LSTMs + KN-5 (equal weights) & 25.3 \\
10 LSTMs + KN-5 (optimal weights) & 25.1 \\
10 LSTMs + SNM10-skip \cite{shazeer2015sparse} & \textbf{23.7} \\
\hline
\end{tabular}
\end{sc}
\end{small}
\end{center}
\vskip -0.1in
\end{table*}

\subsection{Training Procedure}

The models were trained until convergence with an AdaGrad optimizer using a learning rate of 0.2. In all the experiments the RNNs were unrolled for 20 steps without ever resetting the LSTM states. We used a batch size of 128. We clip the gradients of the LSTM weights such that their norm is bounded by 1.0 \cite{pascanu2012difficulty}.

Using these hyper-parameters we found large LSTMs to be relatively easy to train. The same learning rate was used in almost all of the experiments. In a few cases we had to reduce it by an order of magnitude. Unless otherwise stated, the experiments were performed with 32 GPU workers and asynchronous gradient updates. Further details will be fully specified with the code upon publication.

Training a model for such large target vocabulary (793471 words) required to be careful with some details about the approximation to full Softmax using importance sampling. We used a large number of negative (or noise) samples: 8192 such samples were drawn per step, but were shared across all the target words in the batch (2560 total, i.e. 128 times 20 unrolled steps). This results in multiplying (2560 x 1024) times (1024 x (8192+1)) (instead of (2560 x 1024) times (1024 x 793471)), i.e. about 100-fold less computation.

\section{Results and Analysis}
\label{results}

In this section we summarize the results of our experiments and do an in-depth analysis. Table~\ref{main-results-table} contains all results for our models compared to previously published work. Table~\ref{ensemble-results-table} shows previous and our own work on ensembles of models. We hope that our encouraging results, which improved the best perplexity of a single model from 51.3 to 30.0 (whilst reducing the model size considerably), and set a new record with ensembles at 23.7, will enable rapid research and progress to advance Language Modeling. For this purpose, we will release the model weights and recipes upon publication.

\subsection{Size Matters}

Unsurprisingly, size matters: when training on a very large and complex data set, fitting the training data with an LSTM is fairly challenging. Thus, the size of the LSTM layer is a very important factor that influences the results, as seen in Table~\ref{main-results-table}. The best models are the largest we were able to fit into a GPU memory. Our largest model was a 2-layer LSTM with 8192+1024 dimensional recurrent state in each of the layers. Increasing the embedding and projection size also helps but causes a large increase in the number of parameters, which is less desirable. Lastly, training an RNN instead of an LSTM yields poorer results (about 5 perplexity worse) for a comparable model size.

\subsection{Regularization Importance}

As shown in Table~\ref{main-results-table}, using dropout improves the results. To our surprise, even relatively small models (e.g., single layer LSTM with 2048 units projected to 512 dimensional outputs) can over-fit the training set if trained long enough, eventually yielding holdout set degradation.

Using dropout on non-recurrent connections largely mitigates these issues. While over-fitting still occurs, there is no more need for early stopping. For models that had 4096 or less units in the LSTM layer, we used 10\% dropout probability. For larger models, 25\% was significantly better. Even with such regularization, perplexities on the training set can be as much as 6 points below test.

In one experiment we tried to use a smaller vocabulary comprising of the 100,000 most frequent words and found the difference between train and test to be smaller -- which suggests that too much capacity is given to rare words. This is less of an issue with character CNN embedding models as the embeddings are shared across all words.

\subsection{Importance Sampling is Data Efficient}
Table~\ref{nce-vs-sampled} shows the test perplexities of NCE vs IS loss after a few epochs of 2048 unit LSTM with 512 projection. The IS objective significantly improves the speed and the overall performance of the model when compared to NCE.

\begin{table}[t]
\caption{The test perplexities of an LSTM-2048-512 trained with different losses versus number of epochs. The model needs about 40 minutes per epoch. First epoch is a bit slower because we slowly increase the number of workers. }
\label{nce-vs-sampled}
\vskip 0.15in
\begin{center}
\begin{small}
\begin{sc}
\begin{tabular}{lccc}
\hline
\abovespace\belowspace
Epochs & NCE & IS & Training Time [Hours] \\
\hline
\abovespace
1 & 97 & 60 & 1 \\
5 & 58 & 47.5 & 4 \\
10 & 53 & 45 & 8 \\
20 & 49 & 44 & 14 \\
50 & 46.1 & 43.7 & 34 \\
\hline
\end{tabular}
\end{sc}
\end{small}
\end{center}
\vskip -0.1in
\end{table}

\subsection{Word Embeddings vs Character CNN}
Replacing the embedding layer with a parametrized neural network that process characters of a given word allows the model to consume arbitrary words and is not restricted to a fixed vocabulary. This property is useful for data sets with conversational or informal text as well as for morphologically rich languages. Our experiments show that using character-level embeddings is feasible and does not degrade performance -- in fact, our best single model uses a Character CNN embedding.

An additional advantage is that the number of parameters of the input layer is reduced by a factor of 11 (though training speed is slightly worse). For inference, the embeddings can be precomputed so there is no speed penalty. Overall, the embedding of the best model is parametrized by 72M weights (down from 820M weights).

Table~\ref{nearest-neighbors} shows a few examples of nearest neighbor embeddings for some out-of-vocabulary words when character CNNs are used. 

\subsection{Smaller Models with CNN Softmax}

\begin{table}[t]
\caption{Nearest neighbors in the character CNN embedding space of a few out-of-vocabulary words. Even for words that the model has never seen, the model usually still finds reasonable neighbors.}
\label{nearest-neighbors}
\vskip 0.15in
\begin{center}
\begin{tiny}
\begin{sc}
\begin{tabular}{cccc}
\hline
\abovespace\belowspace
Word & Top-1 & Top-2 & Top-3 \\
\hline
\abovespace
incerdible & incredible & nonedible & extendible \\
www.a.com & www.aa.com & www.aaa.com & www.ca.com \\
7546 & 7646 & 7534 & 8566 \\
TownHal1 & TownHall & DJc2 & Moodswing360 \\
Komarski & Koharski & Konarski & Komanski \\
\hline
\end{tabular}
\end{sc}
\end{tiny}
\end{center}
% \vskip -0.1in
\end{table}

Even with character-level embeddings, the model is still fairly large (though much smaller than the best competing models from previous work). Most of the parameters are in the linear layer before the Softmax: 820M versus a total of 1.04B parameters.

In one of the experiments we froze the word-LSTM after convergence and replaced the Softmax layer with the CNN Softmax sub-network. Without any fine-tuning that model was able to reach 39.8 perplexity with only 293M weights (as seen in Table~\ref{main-results-table}).

As described in Section~\ref{cnn_softmax}, adding a ``correction'' word embedding term alleviates the gap between regular and CNN Softmax. Indeed, we can trade-off model size versus perplexity. For instance, by adding 100M weights (through a 128 dimensional bottleneck embedding) we achieve 35.8 perplexity (see Table~\ref{main-results-table}).

To contrast with the CNN Softmax, we also evaluated a model that replaces the Softmax layer with a smaller LSTM that predicts one character at a time (see Section~\ref{charlstm}). Such a model does not have to learn long dependencies because the base LSTM still operates at the word-level (see Figure~\ref{main_figure}(c)). With a single-layer LSTM of 1024 units we reached 49.0 test perplexity, far below the best model. In order to make the comparisons more fair, we performed a very expensive marginalization over the words in the vocabulary (to rule out words not in the dictionary which the character LSTM would assign some probability). When doing this marginalization, the perplexity improved a bit down to 47.9.

\begin{figure}[ht]
\vskip 0.2in
\begin{center}
\centerline{\includegraphics[width=\columnwidth]{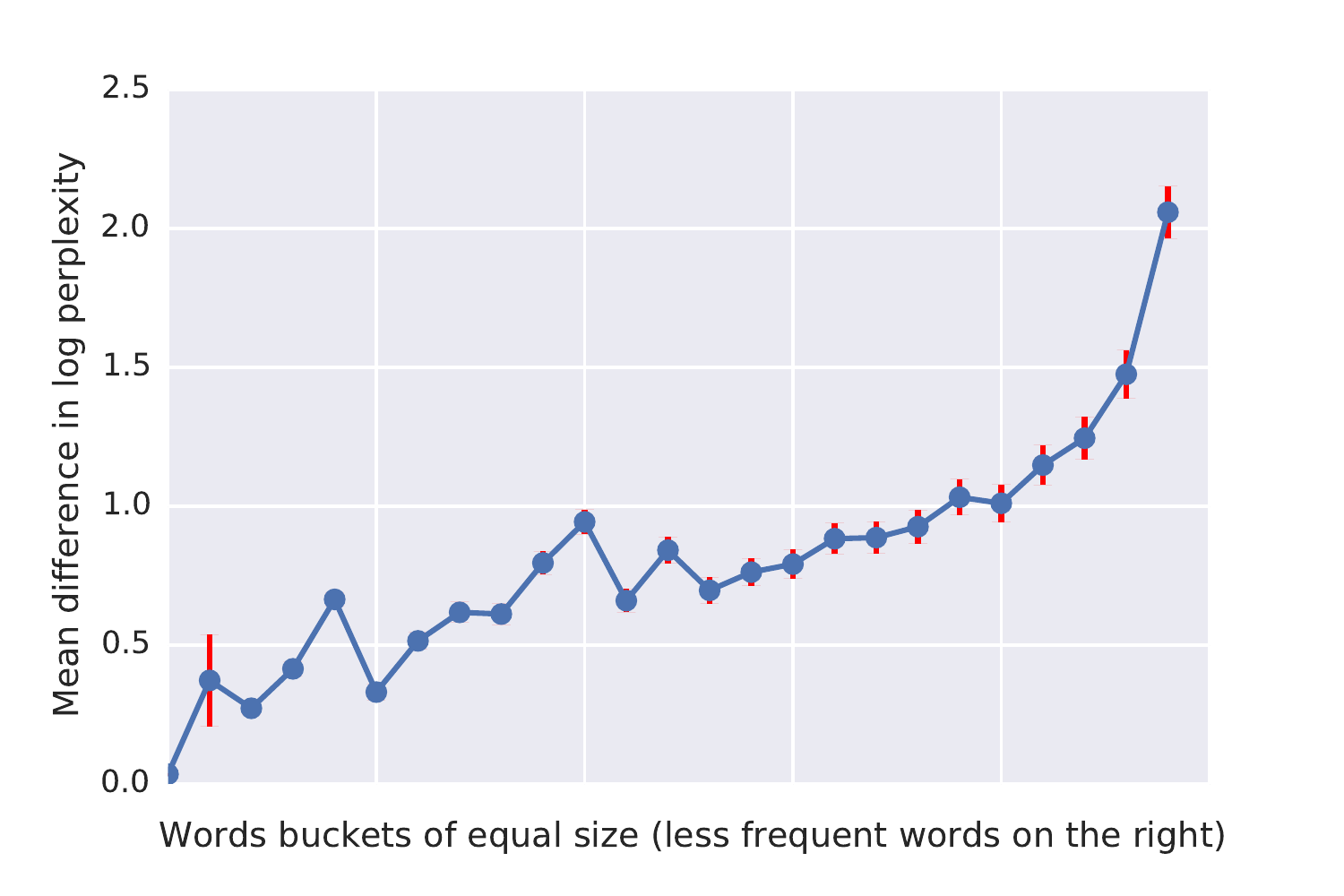}}
\caption{The difference in log probabilities between the best LSTM and KN-5 (higher is better). The words from the holdout set are grouped into 25 buckets of equal size based on their frequencies.}
\label{tail-words}
\end{center}
\vskip -0.2in
\end{figure}

\subsection{Training Speed}

We used 32 Tesla K40 GPUs to train our models. The smaller version of the LSTM model with 2048 units and 512 projections needs less than 10 hours to reach below 45 perplexity and after only \textbf{2 hours} of training the model beats previous state-of-the art on this data set. The best model needs about 5 days to get to 35 perplexity and 10 days to 32.5. The best results were achieved after 3 weeks of training. See Table~\ref{nce-vs-sampled} for more details.

\subsection{Ensembles}

We averaged several of our best models and we were able to reach 23.7 test perplexity (more details and results can be seen in Table~\ref{ensemble-results-table}), which is more than 40\% improvement over previous work. Interestingly, including the best N-gram model reduces the perplexity by 1.2 point even though the model is rather weak on its own (67.6 perplexity). Most previous work had to either ensemble with the best N-gram model (as their RNN only used a limited output vocabulary of a few thousand words), or use N-gram features as additional input to the RNN. Our results, on the contrary, suggest that N-grams are of limited benefit, and suggest that a carefully trained LSTM LM is the most competitive model.

\subsection{LSTMs are best on the tail words}

Figure~\ref{tail-words} shows the difference in log probabilities between our best model (at 30.0 perplexity) and the KN-5. As can be seen from the plot, the LSTM is better across all the buckets and significantly outperforms KN-5 on the rare words. This is encouraging as it seems to suggest that LSTM LMs may fare even better for languages or data sets where the number of rare words is larger than traditional N-gram models.

\subsection{Samples from the model}

To qualitatively evaluate the model, we sampled many sentences. We discarded short and politically incorrect ones, but the sample shown below is otherwise ``raw'' (i.e., not hand picked). The samples are of high quality -- which is not a surprise, given the perplexities attained -- but there are still some occasional mistakes.

Sentences generated by the ensemble (about 26 perplexity):
\begin{quotation} \noindent \scriptsize
$<S>$ With even more new technologies coming onto the market quickly during the past three years , an increasing number of companies now must tackle the ever-changing and ever-changing environmental challenges online .
$<S>$ Check back for updates on this breaking news story .
$<S>$ About 800 people gathered at Hever Castle on Long Beach from noon to 2pm , three to four times that of the funeral cort\`ege .
$<S>$ We are aware of written instructions from the copyright holder not to , in any way , mention Rosenberg 's negative comments if they are relevant as indicated in the documents , " eBay said in a statement .
$<S>$ It is now known that coffee and cacao products can do no harm on the body .
$<S>$ Yuri Zhirkov was in attendance at the Stamford Bridge at the start of the second half but neither Drogba nor Malouda was able to push on through the Barcelona defence .
\end{quotation}

\section{Discussion and Conclusions}
\label{discussion}

In this paper we have shown that RNN LMs can be trained on large amounts of data, and outperform competing models including carefully tuned N-grams. The reduction in perplexity from 51.3 to 30.0 is due to several key components which we studied in this paper. Thus, a large, regularized LSTM LM, with projection layers and trained with an approximation to the true Softmax with importance sampling performs much better than N-grams. Unlike previous work, we do not require to interpolate both the RNN LM and the N-gram, and the gains of doing so are rather marginal.

By exploring recent advances in model architectures (e.g. LSTMs), exploiting small character CNNs, and by sharing our findings in this paper and accompanying code and models (to be released upon publication), we hope to inspire research on large scale Language Modeling, a problem we consider crucial towards language understanding. We hope for future research to focus on reasonably sized datasets taking inspiration from recent advances seen in the computer vision community thanks to efforts such as Imagenet \cite{deng2009imagenet}.

\section*{Acknowledgements} 

We thank Ciprian Chelba, Ilya Sutskever, and the Google Brain Team for their help and discussions. We also thank Koray Kavukcuoglu for his help with the manuscript.

\bibliography{lm_paper}
\bibliographystyle{icml2016}

\end{document}